\documentclass{article}

\usepackage[preprint]{neurips_2022}

\usepackage[utf8]{inputenc} 
\usepackage[T1]{fontenc}    
\usepackage{hyperref}       
\usepackage{url}            
\usepackage{booktabs}       
\usepackage{amsfonts}       
\usepackage{nicefrac}       
\usepackage{microtype}      
\usepackage{xcolor}         
\usepackage{array,graphicx}
\usepackage{soul}
\usepackage{float}
\usepackage{caption}
\usepackage{wrapfig}
\usepackage{comment}
\usepackage[compact]{titlesec}
\titlespacing{\section}{0pt}{2ex}{1ex}
\titlespacing{\subsection}{0pt}{1ex}{0ex}

\usepackage[noend]{algpseudocode}
\usepackage[linesnumbered,ruled,noline,procnumbered,noend]{algorithm2e}
\newenvironment{protocol}[1][t]
  {
   \begin{algorithm}[#1]%
   \scriptsize{ }
  }{\end{algorithm}}
\SetAlCapNameFnt{\small}
\SetAlCapFnt{\small}
\usepackage{amsthm}

\usepackage{enumitem}
\usepackage{multirow}
\usepackage[normalem]{ulem}
\usepackage{amsmath}
\usepackage{pifont}
\usepackage{bm}
\usepackage{xspace}

\usepackage{color}

\usepackage{soul}

\DeclareMathOperator{\sgn}{sgn}

\newcommand{\pRDM}{\ensuremath{\pi_{\mathsf{GR-RANDOM}}}\xspace}
\newcommand{\pDIV}{\ensuremath{\pi_{\mathsf{DIV}}}\xspace}
\newcommand{\pMUL}{\ensuremath{\pi_{\mathsf{MUL}}}\xspace}

\newcommand{\pLN}{\ensuremath{\pi_{\mathsf{LN}}}\xspace}
\newcommand{\pEQ}{\ensuremath{\pi_{\mathsf{EQ}}}\xspace}
\newcommand{\pGTE}{\ensuremath{\pi_{\mathsf{GTE}}}\xspace}

\newcommand{\pPRE}{\ensuremath{\pi_{\mathsf{RW}}}\xspace}
\newcommand{\pLAP}{\ensuremath{\pi_{\mathsf{LAP}}}\xspace}

\newcommand{\pPOST}{\ensuremath{\pi_{\mathsf{ROC}}}\xspace}

\newcommand{\pCF}{\ensuremath{\pi_{\mathsf{CF}}}\xspace}

\newcommand{\xmark}{\ding{55}}%

\usepackage{amssymb}

\newcommand\extrafootertext[1]{%
    \bgroup
    \renewcommand\thefootnote{\fnsymbol{footnote}}%
    \renewcommand\thempfootnote{\fnsymbol{mpfootnote}}%
    \footnotetext[0]{#1}%
    \egroup
}

\title{PrivFairFL: Privacy-Preserving\\
Group Fairness in Federated Learning}

\author{%
  Sikha Pentyala\textsuperscript{\rm \textdagger,\S}, 
  Nicola Neophytou\textsuperscript{\rm \textdagger,\textdaggerdbl}, 
  Anderson Nascimento \textsuperscript{\rm \S}\\
  \textbf{Martine De Cock} \textsuperscript{\rm \S,\#}
  \textbf{Golnoosh Farnadi}\textsuperscript{\rm \textdagger,\textdaggerdbl} 
  \thanks{\textsuperscript{\rm \textdagger} Mila -- Quebec AI Institute;\textsuperscript{\rm \textdaggerdbl} Department of Decision Sciences, HEC Montr\'eal;\textsuperscript{\rm \S} School of Engineering and Technology, University of Washington Tacoma;   \textsuperscript{\rm \#} Department of Applied Mathematics, Computer Science and Statistics, Ghent University; Correspondence to Sikha Pentyala (sikha@uw.edu) }
}

\begin{document}

\maketitle

\begin{abstract}
Group fairness ensures that the outcome of machine learning (ML) based decision making systems are not biased towards a certain group of people defined by  a sensitive attribute such as gender or ethnicity. Achieving group fairness in Federated Learning (FL) is challenging because mitigating bias inherently requires using the sensitive attribute values of all clients, while FL is aimed precisely at protecting privacy by \textit{not} giving access to the clients' data. As we show in this paper, this conflict between fairness and privacy in FL can be resolved by combining FL with Secure Multiparty Computation (MPC) and Differential Privacy (DP). In doing so, we propose a method for training group-fair ML models in cross-device FL under complete and formal privacy guarantees, without requiring the clients to disclose their sensitive attribute values.
Empirical evaluations on real world datasets demonstrate the effectiveness of our solution to train fair and accurate ML models in federated cross-device setups with privacy guarantees to the users.

\end{abstract}


\section{Introduction}\label{SEC:intro}

Machine learning (ML) models are widely adopted in decision making systems, directly affecting quality of life, including in healthcare, justice, education, surveillance, human resources, and advertising. 
The use of ML in such impactful domains has raised valid concerns regarding fairness of models that make biased predictions, resulting in discrimination of individuals, disparity in allocation of resources, and inequity in quality of service. Discrimination by ML models has been reported in applications for  recidivism prediction \cite{angwin2016machine,larson2016we}, credit card approval \cite{vigdor2019apple}, advertising \cite{sweeney2013discrimination,ali2019discrimination}, and job matching \cite{linkedin:2021}, among others \cite{buolamwini2018gender}. 
\textit{Group fairness}, a widely considered notion of fairness that has been  imposed by regulatory bodies in AI applications \cite{eCFR}, 
aims at ensuring that the ML model predictions are not biased towards a certain group of people, as defined by a sensitive attribute such as race, gender, or age  \cite{verma2018fairness}. Developing group-fair ML models in the centralized learning setup where a single entity has access to all data has been well studied in the literature \cite{dwork2012fairness,hardt2016equality,pessach2020algorithmic}. In a lot of applications however, data originates from many different clients who -- out of privacy concerns -- may not wish to, or may legally not be allowed to, disclose their data to a central entity.

Federated learning (FL) \cite{mcmahan2017communication}, a collaborative training paradigm where each client performs training on its own data and shares only model parameters or gradients instead of raw data with a central aggregator, is gaining popularity as a privacy-enhancing technology (PET) \cite{gboard,kairouz2021advances,paulik2021federated,google}. 
Many FL applications naturally exhibit a \textit{cross-device  federated setup} with horizontally partitioned data, 
where each client in the federation has one particular value for the sensitive attribute (e.g.~a female client, using a mobile recommending application, will have user-item interaction data belonging only to the female group).  Methods for training group-fair ML models in the \textit{centralized} paradigm rely on knowledge of sensitive attribute values across the entire dataset, for instance to compute aggregated statistics for data rebalancing, or to compute a bias correction term in the loss function. Extending these methods to FL is non-trivial. \textit{Fair model training in FL} has been identified as challenging due to the intrinsic conflict between bias mitigation algorithms' need to perform computations over the data of all clients, and FL's aims to preserve data privacy by \textit{not} giving access to client data \cite{zhou2021towards}. The emerging literature on group-fair FL works around this by assuming a \textit{cross-silo} set-up in which each client has data for \textit{multiple} sensitive attribute values (e.g.~each client represents a bank with data about female as well as male customers)  
\cite{mohri2019agnostic,abay2020mitigating,du2021fairness,fcfl} and/or by
sacrificing some privacy for fairness. 
Recently proposed methods  (\cite{fedfairFace,papadaki2021federating,yue2021gifair}) send unprotected model parameters and/or other unprotected information such as fairness metrics to a central aggregator, thereby leaking 
information about clients' data \cite{boenisch2021curious}. 
FairFed \cite{ezzeldin2021fairfed} 
employs techniques 
to protect the individual updates but the aggregations at the central aggregator are not protected 
and can reveal information about the clients' data when under attack. 
In this paper, to the best of our knowledge, we propose the first method for training group-fair ML models in cross-device FL under complete and formal privacy guarantees,
protecting both the sent updates or values as well as the aggregated information. 

\textbf{Our approach.}
Building on the literature on fair ML in the centralized paradigm 
\cite{kamiran2012data,hardt2016equality,faircb}  we propose a \textit{privacy-preserving pre-processing} technique for bias mitigation through 
training sample reweighing ({\tt PrivFairFL-Pre}),
and a \textit{privacy-preserving post-processing} technique that identifies fair classification thresholds for different groups ({\tt PrivFairFL-Post}). The core of these techniques are Secure Multiparty Computation (MPC) protocols for collecting aggregated statistics of the label and sensitive attribute value distributions across the federation. Unlike a traditional central aggregator in FL, the \textit{computing servers} that execute these MPC protocols never see the personal values of the clients in an unencrypted manner.
We further protect any output of such computations that needs to be made public through perturbations 
to provide Differential Privacy (DP) guarantees. To do so, the computing servers use MPC to generate the necessary Laplacian noise and add it to the outputs of the computations to satisfy DP requirements, thereby having an MPC protocol effectively play the role of a trusted curator implementing global DP.
As such, our solution combines the best of multiple PETs, namely FL, MPC, and DP, to train fair and high-utility ML models, as we show through an empirical analysis.
%
%
Our main novel contributions are:
\vspace{-4pt}
\begin{itemize}[leftmargin=*,noitemsep,topsep=0pt]
\item MPC protocols for (1) collecting ROC curves and statistics of label and sensitive attribute value distributions across a federation of clients in a fully privacy-preserving manner, and for (2) the publication of said ROC curves and statistics under DP guarantees.
\item Pre- and post-processing based algorithms {\tt PrivFairFL-Pre} and {\tt PrivFairFL-Post} that leverage these statistics for training group-fair ML models in the FL paradigm, without requiring the clients to disclose the values of their sensitive attributes. 
\item An empirical evaluation of our proposed {\tt PrivFairFL} methods 
to demonstrate that group fairness can be achieved in cross-device FL without leaking sensitive user information. 
\end{itemize}

\section{Preliminaries}\label{SEC:prelim}

\textbf{Notations.}\label{prelim:notn}
We consider a federation with $M$ clients, in which each client $k$ ($k = 1 \ldots M$) holds a dataset $D_k$ with $n_k$ training samples. <$X_{ik},s_{ik},y_{ik}$> represents the $i^{th}$ training sample held by the $k^{th}$ client where $s_{ik}$ is the value of a sensitive attribute, $y_{ik}$ is the value of a class label, and $X_{ik}$ is the set of remaining feature values. $N = \sum_{k=1}^{M} n_k$ denotes the total number of samples. 
$S$ denotes the sensitive attribute and $Y$ denotes the class label. For the sake of simplicity, we focus on the case where $S$ represents a binary sensitive attribute that takes $u$ (representing the unprivileged group) and $p$ (representing the privileged group) as its values, and $Y$ takes $0$ and $1$ as its values. Our proposed methods, however, can be extended to the case of multiple sensitive attributes, including non-binary, and multi-class classification (see Sec.~\ref{SEC:extension}).
We use $C(s,y)$ to denote the count of all training samples from all the clients with sensitive attribute value $s$ and class label $y$:
\begin{equation}\label{eqcounts}
    C(s,y) = \sum\limits_{k=1}^M \sum\limits_{i=1}^{n_k} \# \{\mbox{<}X_{ik},s_{ik},y_{ik}\mbox{>} \,|\, s_{ik} = s \wedge y_{ik} = y\}
\end{equation}
For example, $C(u,1)$ is the number of training samples from all clients belonging to the unprotected group and with class label $1$. The model parameters of client $k$ are denoted as $\theta_k$ and the aggregated model parameters are represented by $\theta$. The model $\theta$ makes predictions $\Hat{Y}$. 




\looseness = -1
\textbf{Federated Learning (FL).}\label{prelim:fl}
FL is a collaborative learning paradigm where the clients 
together train an ML model in coordination with a central server
while keeping the training data private and decentralized. The clients train the model on their own data locally
and send only the gradients or model parameters $\theta_k$ to the central server for aggregation. This is done iteratively for a defined number of rounds to get a global model $\theta$ that has achieved the learning objective on the combined training data of the clients. This enables the clients to keep their data within the premises providing first-level privacy to the clients. It is well-known that FL alone does not offer formal privacy guarantees, and that it leaves the clients vulnerable to information leakage \cite{kairouz2021advances}. MPC-based methods \cite{bonawitz2017practical} and DP-based methods \cite{abadi2016deep} have been proposed to be used in combination with FL to provide stronger privacy guarantees. 

\looseness = -1
\textbf{Differential Privacy (DP).}\label{prelim:dp}
DP guarantees plausible deniability regarding an instance being in a dataset, 
hence offering privacy guarantees. DP techniques ensure that a randomized algorithm $\mathcal{A}$ behaves similarly on a dataset $D$ and a neighboring dataset $D'$ that differs in a single entry,\footnote{Throughout this paper we consider event-level DP, i.e.~an entry or instance corresponds to one training example of one client in the federation, and $D$ holds the data of all clients combined, i.e.~$D = \cup_{k=1}^M D_k$.} i.e. $\mathcal{A}$ generates a similar output probability distribution on $D$ and $D'$ \cite{dwork2014algorithmic}. $\mathcal{A}$ can for instance be an algorithm that takes as input a dataset $D$ of training examples and outputs an ML model; alternatively $\mathcal{A}$ can return counts or other statistics about $D$.
A randomized algorithm $\mathcal{A}$ is called $(\epsilon, \delta)$-DP if for all pairs of neighboring sets $D$ and $D'$, and for all subsets $S$ of $\mathcal{A}$'s range,
\vspace{-3pt}
\begin{equation}\label{DEF:DP}
\mbox{P}(\mathcal{A}(D) \in S) \leq e^{\epsilon} \cdot \mbox{P}(\mathcal{A}(D') \in S) + \delta.
\end{equation}
where $\epsilon$ is the privacy budget or privacy loss and $\delta$ is the probability of violation of privacy. 
The smaller these values, the stronger the privacy guarantees. When $\delta = 0$, then $\mathcal{A}$ is said to be $\epsilon$-DP. 
An $(\epsilon,\delta)$-DP randomized algorithm $\mathcal{A}$ is commonly created out of an algorithm $\mathcal{A^*}$ by adding noise that is proportional to the \textit{sensitivity} of $\mathcal{A^*}$, in which the sensitivity measures
the maximum impact a change in the underlying dataset can have on the output of $\mathcal{A^*}$. The post-processing property of DP guarantees that if $\mathcal{A}$ is $\epsilon$-DP, then $g(\mathcal{A})$ is also $\epsilon$-DP where $g$ is an arbitrary function. In other words, 
any arbitrary computations performed on DP output preserves DP without any effect on the privacy budget $\epsilon$. 



The traditional DP paradigm -- \textit{global DP} -- assumes that the entire dataset resides with a central curator who computes the noise. Alternatively, if the data originates from multiple data holders, then they can add noise before sending their information to a trusted curator for further processing. Approaches based on this \textit{local DP} paradigm tend to have lower utility while of course offering better privacy than requiring data holders to disclose their information in an unprotected manner to a trusted curator. In {\tt PrivFairFL} (Sec.~\ref{SEC:methods}) we replace the trusted curator by an MPC protocol that can be run on untrusted servers, achieving the same utility as in the global DP paradigm but without requiring the clients to disclose their sensitive attribute values to anyone.

\textbf{Secure Multiparty Computation (MPC).}\label{prelim:mpc}
MPC is an umbrella term for cryptographic approaches that allow two or more parties (servers) to jointly compute a specified output from their private information in a distributed fashion, without revealing anything beyond the output of the computation to each other \cite{damgard}. MPC protocols are designed to prevent and detect attacks by an adversary $A$ who can corrupt one or more parties to learn the private information or to cause incorrect computations. These protocols can be mathematically proven to guarantee privacy and correctness. We follow the universal composition theorem 
that allows modular design where the protocols remain secure even if composed with other or the same MPC protocols 
\cite{canetti2000security}. For details, see Evans et al.~\cite{evans2018pragmatic}.


All computations in MPC are commonly done on integers modulo $q$, i.e.,~in the ring $\mathbb{Z}_q$. 
In {\tt PrivFairFL},
the clients first convert any real-valued inputs into a fixed point representation\footnote{Real values are represented using $l$ bits in total with $d$ bits for decimal precision.}
and then encrypt their private data (all integers) by splitting them into so-called secret-shares. These secret-shares are then distributed to a set of servers (computing parties) that run MPC protocols and perform computations over these secret shares.  Proper design of the MPC protocols ensures that no server learns anything about the input on its own and that nothing about the input is revealed to any subset of  colluding (malicious) servers that can be corrupted. For example, in the replicated secret-sharing scheme with 3 servers 
by Araki et al.~\cite{araki2016high}, a private value $x$ $\in$ $\mathbb{Z}_{2^f}$ is secret-shared among computing parties $P_1, P_2,$ and $P_3$ by picking uniformly random numbers (shares) $x_1, x_2, x_3 \in \mathbb{Z}_{2^f}$ such that 
$x_1 + x_2 +x_3 =  x \mod{2^f}$,
and distributing $(x_1,x_2)$ to $P_1$, $(x_2,x_3)$ to $P_2$, and $(x_3,x_1)$ to $P_3$. Note that, while the secret-shared information can be trivially revealed by combining shares from any two servers, no single server can obtain any information about $x$ given its shares. In the remainder of this paper, we use $[\![x]\!]$ as a shorthand for a secret sharing of $x$, regardless of which secret-sharing scheme is used. In the replicated secret-sharing scheme sketched above, $[\![x]\!] = ((x_1,x_2),(x_2,x_3),(x_3,x_1))$. 


A variety of MPC schemes are available in the literature that can be employed based on different security threat models (see Tab.~\ref{tab:mpcschemes}). These are defined based on either the number of computing parties that can be corrupted or the modified behavior of the corrupted parties. MPC schemes for the \textit{dishonest-majority} setting offer security even if half or more number of computing parties are corrupted, while MPC schemes for the \textit{honest-majority} setting offer security as long as more than half of the computing parties remain uncorrupted. If the corrupted computing parties adhere to the MPC protocols but try to gather additional information, they are said to be corrupted by a \textit{passive} adversary, whereas if the corrupted computing parties deviate from the protocol instructions, they are said to be corrupted by a \textit{active} adversary. We design our MPC protocols in Sec.~\ref{SEC:methods} building on the primitive MPC protocols from the schemes in Tab.~\ref{tab:mpcschemes}. 
Our protocols are sufficiently generic to be used in dishonest-majority as well as honest-majority settings, with passive or active adversaries. This is achieved by changing the underlying MPC scheme to align with the desired security setting. 

The MPC schemes in Tab.~\ref{tab:mpcschemes} above provide a mechanism for the servers to perform cryptographic primitives through the use of secret shares, namely addition of a constant, multiplication by a constant, and addition of secret-shared values (which can all be done by the servers through carrying out local computations on their own shares) and multiplication of secret-shared values (which requires communication among the servers and is denoted as MPC protocol $\pMUL$ in this paper). Building on these cryptographic primitives, MPC protocols for other operations have been developed. We use a secure division protocol $\pDIV$ with an iterative algorithm that is well known in the MPC literature \cite{catrina2010secure}). We also use a secure comparison protocol $\pGTE$: at the start of $\pGTE$, the parties have secret sharings $[\![x]\!]$ and $[\![y]\!]$ of integers $x$ and $y$; at the end of the protocol they have a secret sharing of $1$ if $x \geq y$, and a secret sharing of $0$ otherwise \cite{mpspdz}. The protocol $\pEQ$ for equality test works similarly.

\looseness = -1
\textbf{Group Fairness.}\label{prelim:fair}
Group fairness measures how balanced the predicted outcomes are across the groups defined by the sensitive attribute $S$. 
Many different notions of group fairness have been proposed in the literature \cite{dwork2012fairness,feldman2015certifying,hardt2016equality,kleinberg2016inherent,donini2018empirical}. We consider popular statistical notions of fairness 
that rely on computing the
true positive rate ({$\mbox{TPR}_p$} and $\mbox{TPR}_u$), 
false positive rate ($\mbox{FPR}_p$ and $\mbox{FPR}_u$),
and the number of true positives ($\mbox{TP}_p$ and $\mbox{TP}_u$) for the privileged ($p$) and unprivileged groups ($u$) respectively. 
%
\textit{Disparate Impact (DI)}  measures discrimination in the predictions by computing the recall (TPR) for the groups \cite{speicher2018potential}. A value \mbox{DI} $=1$ indicates discrimination-free predictions. We report the degree of discrimination using $|1-\mbox{DI}|$. 
%
\textit{Equal opportunity (EOP)} considers the fairness of the predictions from the perspective of TPR 
\cite{hardt2016equality}. This metric focuses only on the positive or advantaged outcome. We report this metric as equal opportunity difference ($\Delta$EOP) by computing the difference between the TPR of the two groups.
\textit{Equalized odds (EODD)} considers the predictions fair if $\hat{Y}$ and $S$ are independent conditional on $Y$ \cite{hardt2016equality}. This implies that both TPR and FPR  are equal between the groups. 
We report this metric as average odds difference ($\Delta$EODD). 
%
\textit{Statistical parity (SP)} considers the predictions as fair if the number of positive predictions are the same for the two groups 
\cite{dwork2012fairness}. We report statistical parity difference ($\Delta$SP) as the difference between the ratio of positive outcomes per group.
Formally:
\begin{eqnarray}
|1-\mbox{DI}| & = &   |1-\max(\mbox{TPR}_u / \mbox{TPR}_p, \mbox{TPR}_p / \mbox{TPR}_u)|\\
\Delta\mbox{EOP} & = &  |\mbox{TPR}_p - \mbox{TPR}_u| \\
\Delta\mbox{EODD} & = &   0.5 \cdot ( |\mbox{TPR}_p - \mbox{TPR}_u| + |\mbox{TPR}_p - \mbox{TPR}_u|) \\
\Delta\mbox{SP} & = &  |(\mbox{TP}_p/\mbox{N}_p) - (\mbox{TP}_u/\mbox{N}_u)|
\end{eqnarray}
where $\mbox{N}_p$ and $\mbox{N}_u$ denote the number of samples for privileged and unprivileged group respectively. The lower the values of the metrics, the fairer the predictions made by the model.





\section{Related Work}\label{SEC:relatedw}
\vspace{-3pt}
\noindent
\textbf{Group fairness in centralized learning.}\label{RL:mitigation}
Existing unfairness mitigation techniques can be categorized into three categories based on the stage in the ML pipeline they are incorporated into.
\textit{Pre-processing techniques} are applied to the training data to create a less biased dataset \cite{kamiran2012data,zemel2013learning, feldman2015certifying,calmon2017optimized,faircb}. They either modify the raw data by changing the sensitive attribute and/or the class label, or assign weights to the samples based on their label and/or sensitive attribute. 
Reweighing techniques make the loss function penalize incorrect predictions based on the assigned weights of the samples to learn a fair predictor across groups. \textit{In-processing techniques} are employed during the training phase of a model \cite{kamishima2012fairness, zhang2018mitigating, agarwal2018reductions,kearns2018preventing,celis2019classification}. They usually modify the optimization problem by either adding a regularizer to the objective function or constraints to the optimization formulation. 
\textit{Post-processing techniques}  are applied to the predicted labels to generate fairer predictions either by flipping the labels or finding an optimal threshold \cite{kamiran2012decision,hardt2016equality}. Unlike in-processing,  the pre- and post-processing techniques are independent of the notions of fairness, learning objective, and the model being trained. 
{\tt PrivFairFL} extends pre- and post-processing to FL.



\vspace{-1pt}
\noindent
\textbf{Group fairness in FL.}\label{RL:fpfl}
Extending bias mitigation techniques from the centralized paradigm to FL is challenging due to an intrinsic conflict between fair model training and FL \cite{zhou2021towards}:
(i) evaluating the fairness of a model, or mitigating bias, requires access to the data of all clients; and (ii) 
FL aims at preserving data privacy by \textit{not} giving such access. 
\looseness = -1
As Tab.~\ref{TAB:relatedwork}
(* See Sec.\ref{SEC:extension})
shows, methods for training group-fair models have been proposed for cross-silo setups (each client has data for multiple sensitive attribute values), and for cross-device setups (each client has data for only one sensitive attribute value). Nearly all methods are based on in-processing, hence tailored to a specific training algorithm, model architecture, and fairness notion \cite{fcfl,ugffl,du2021fairness}. Furthermore, as the ``privacy'' column in Tab.~\ref{TAB:relatedwork} indicates, most of the emerging literature on group-fair FL tries to work around the conflict between (i) and (ii) by sacrificing privacy for fairness.
Information leaks can occur when the clients send updated model parameters, gradients, fairness metrics, or the values of sensitive attributes to the aggregator, or by analyzing the aggregated outputs.  For example, during FL training, assuming that an adversary $A$ has the model from the previous round and the gradient updates from the current round, $A$ can infer a private training example \cite{kairouz2021advances}. Current works  
do not take into account such information leaks in FL \cite{papadaki2021federating,yue2021gifair,fedfairFace,hong2021federated}. $A$ 
can also analyze the aggregated outputs to infer knowledge about a particular client. 
FairFed \cite{ezzeldin2021fairfed}  and Rodriguez et al. \cite{rodriguez2021enforcing} employ SecAgg \cite{bonawitz2017practical} to protect data leaks from information sent by the clients, but fail to protect the aggregated values. Though Rodriguez et al.~\cite{rodriguez2021enforcing} use a combination of MPC and DP, they reveal the aggregated gradients to the computing parties and add noise in-the-clear to publish DP aggregates to the FL aggregator. Similarly, Zhang et al.~\cite{zhang2020fairfl} protect only the discrimination indices sent by clients and fail to protect any aggregated output and gradient updates.  

Our proposed {\tt PrivFairFL} methods provide provable privacy guarantees to the clients by employing techniques from MPC and DP during training and for aggregation, to protect both the input (training data and/or other sensitive attributes of the clients) and the output (the aggregations and the global model). The closest to our work is that of Abay et al.~\cite{abay2020mitigating} who privately debias the training data by reweighing the training samples as a pre-processing mitigation technique and use local DP to protect the values of sensitive attributes. As per their analysis, their proposed technique results in non-uniform effects on group fairness metrics over different datasets for cross-device setups and performs well only for cross-silo setups. We include this local DP preprocessing approach as a baseline method in Tab.~\ref{tab:results}, and find that it is indeed outperformed by {\tt PrivFairFL-Pre}. 



\vspace{-1pt}
\noindent
\textbf{Private aggregation techniques.}\label{RL:fpfl}
Statistics about the underlying data distribution are commonly used to improve the ML model learning process.
Various works have shown that statistics about the data distribution across clients in FL can improve the utility of models or make them more fair, especially for clients with imbalanced or non-i.i.d.~data \cite{duan2019astraea,du2021fairness}. The challenge is to collect the statistics without infringing upon the clients' privacy.
Solutions for aggregation with MPC  to protect the private input data 
have been recently employed in FL to train group-fair models \cite{ezzeldin2021fairfed,zhang2020fairfl}. However, the output of such MPC-based aggregations can still leak information about the client's data. 
In {\tt PrivFairFL} we go a step further by adding noise to the aggregated values to provide DP guarantees. 
Existing works for aggregation do so by having the clients participate in the noise generation. Such solutions are not resilient to malicious clients, and require extensive communication between clients and the aggregator  \cite{acsdream,DPaggstar}. Our proposal correctly generates the noise in a secure way, inside MPC protocols that are resilient to corruptions by semi-honest and malicious adversaries 
(see Sec.~\ref{SEC:methods}). 

\begin{table}[H]
\captionof{table}{Related work on Group Fairness in FL
\label{TAB:relatedwork}}

\begin{tabular}{l cc  cc ccc}
\toprule
\textbf{Paper} & \multicolumn{2}{c}{\textbf{Scenario}} & 
\multicolumn{2}{c}{\textbf{Privacy}} & \multicolumn{3}{c}{\textbf{Mitigation Alg.}}\\ 
& silo & dev & DP & MPC & Pre & In & Post\\
\midrule

Abay~\cite{abay2020mitigating} & \checkmark &  & 
\checkmark & \xmark & \checkmark & 
\checkmark & \xmark \\ 

AgnosticFair \cite{du2021fairness} & 
\checkmark & & 
\xmark & \xmark & \xmark & \checkmark & \xmark\\ 

FCFL \cite{fcfl} & \checkmark & & 
\xmark & \xmark & \xmark & \checkmark & \xmark\\ 

Zhang~\cite{ugffl} & \checkmark & & 
\xmark & \xmark & \xmark & \checkmark & \xmark\\ 

FairFL \cite{zhang2020fairfl} & \checkmark & & 
\xmark & \checkmark & \xmark & \checkmark & \xmark\\ 

FPFL \cite{padala2021federated} & \checkmark & & 
\checkmark & \xmark & \xmark & \checkmark & \xmark\\ 

Rodriguez~\cite{rodriguez2021enforcing} & \checkmark & \xmark & 
\checkmark & \checkmark & \xmark & \checkmark & \xmark \\ 

\midrule

Kanaparthy~\cite{fedfairFace}
&  & \checkmark 
& 
\xmark & \xmark
& \xmark & \checkmark & \xmark \\

GI-FAIR \cite{yue2021gifair} &  & \checkmark & 
\xmark & \xmark & \xmark & \checkmark & \xmark\\

FADE \cite{hong2021federated} &  & \checkmark & 
\xmark & \xmark & \xmark & \checkmark & \xmark\\

\midrule
FairFed \cite{ezzeldin2021fairfed} & \checkmark & \checkmark & 
\xmark & \checkmark & \xmark & \checkmark & \xmark\\

Papadaki~\cite{papadaki2021federating} & \checkmark & \checkmark & 
\xmark & \xmark & \xmark & \checkmark & \xmark \\ 

\midrule
\textbf{PrivFairFL} &  * & \checkmark & 
\checkmark &\checkmark & \checkmark & * & \checkmark \\
\bottomrule
\end{tabular}

\end{table}

\begin{protocol}[H]
   \SetKwInOut{Input}{Input}
    \SetKwInOut{Output}{Output}
    \Input{Scale $b$}
    \Output{Secret-shared value $[\![x]\!]$ drawn from Laplace distribution with mean $0$ and scale $b$}
    {
            
            $[\![u]\!] \leftarrow \pRDM(-0.5,0.5)$\\    
            
            $[\![\mbox{sgn}_u]\!] \leftarrow \pGTE([\![u]\!],0)$\\ 
            $[\![\mbox{abs}_u]\!] \leftarrow \pMUL([\![u]\!],[\![\mbox{sgn}_u]\!]))$ \\      
            $[\![\mbox{ln}_u]\!] \leftarrow \pLN(1 - 2 \cdot [\![\mbox{abs}_u]\!])$\\

            $[\![x]\!] \leftarrow -b \cdot \pMUL([\![\mbox{ln}_u]\!], [\![\mbox{sgn}_u]\!])$\\

   }
    \KwRet{$[\![x]\!]$}
    \caption{$\pLAP$ for secure sampling from Laplacian distribution}\label{prot:lap}
\end{protocol}

\section{Methodology}\label{SEC:methods}
\textbf{Computational setting.} There are $M$ clients who each have a dataset $D_k$ as described in Sec.~\ref{SEC:prelim}. There are $r \geq 2$ computing parties (servers) who can execute MPC protocols and perform other computations to aid the clients in model training. We propose two strategies for bias mitigation, namely one that is applied before model training ({\tt PrivFairFL-Pre}) and one that is applied after model training ({\tt PrivFairFL-Post}). Both strategies are independent of the model training phase, which means that they can be combined with any technique for model training in FL, including DP-SGD \cite{abadi2016deep}. While all $r$ servers participate in the MPC protocols for bias mitigation, the aggregation of the weights or gradients during model training can either be performed by one of the servers -- which then acts like the traditional central aggregator in traditional FL -- or it can be implemented as a straightforward MPC protocol ran by all the servers. The methods described in this section are generic and work with either aggregation setup during the model training phase.
\looseness=-1
We propose two strategies for training group-fair models in FL with formal privacy guarantees. Each strategy is based on the clients in the federation sending encrypted shares of their information to the $r$ computing parties in an ``MPC as a service'' setting. These parties (1) run MPC protocols to perform computations for bias mitigation, including an MPC subprotocol to add secret-shared noise to the secret-shared results of those computations to provide the desired DP guarantees, and (2) publish the outcomes of this process back to the clients. The two strategies differ in when the MPC protocols are executed (before or after model training), what the input and output is, and what computations are being performed. Performing bias mitigation inside MPC protocols yields the same utility and fairness as one could achieve with global DP, with the added advantage that the clients do not need to disclose their data to a trusted curator.

\subsection{Privacy Preserving Pre-processing for Bias Mitigation}\label{SEC:preproc}
In {\tt PrivFairFL-Pre}, the unified training dataset $D$ is debiased 
by assigning weights to the samples based on the values of $S$ and/or $Y$. Such global assignment of weights on the unified data addresses the heterogeneous data distributions in cross-device setups. A reweighing technique from the centralized learning domain is to assign a training instance with sensitive attribute value $s$ and label $y$ a weight of $1 / C(s,y)$, with $C(s,y)$ the total number of examples in the training data with that sensitive attribute and label value \cite{fairbalance}. We adopt this idea in protocol $\pPRE$, and explain below how $\pPRE$ can be extended to other weight balancing techniques as well.

In FL, each client $k$ has its local dataset $D_k$, and needs to obtain weights for the instances in $D_k$ that are based on counts $C(s,y)$ across the entire federation. To obtain these weights, each client $k$ starts by counting the number of negative and positive instances in $D_k$. We denote these local counts as $LC(0,k)$ and $LC(1,k)$ respectively, for $k=1 \ldots M$. Next each client encrypts this information by splitting it into secret shares and sending it to the computing parties $P_i$ ($i=1 \ldots r$),\footnote{We implemented our solutions for $r=2, 3$ and $4$ (see Sec.~\ref{SEC:results}) but they are general and work with any number of computing parties.} along with secret shares of $T(k)$ which denotes whether client $k$ belongs to the protected group $T(k) = 1$ or to the unprotected group $T(k)=0$. None of the computing parties (servers) can derive any information about the local counts or sensitive attribute values from the secret shares received.

\begin{minipage}[t]{0.51\textwidth}
\begin{protocol}[H]
   \SetKwInOut{Input}{Input}
    \SetKwInOut{Output}{Output}
    \Input{Total number $M$ of clients; 
    secret-shares of vector $T$ of length $M$ denoting whether client $k$ belongs to the protected group ($T(k) = 1$) or to the unprotected group ($T(k) = 0$);
    secret-shares of $2 \times M$ matrix $LC$ with local counts  of number of negative instances $LC(0,k)$ and positive instances $LC(1,k)$ for client $k$;
    $\epsilon$ privacy budget allotted for bias mitigation}
    \Output{Secret-shares of sample weights for negative and positive instances in protected and unprotected groups}
    
    {
    
    \For{$s$ \rm{in} $\{p,u\}$ \rm{and} $y$ \rm{in} $\{0,1\}$}{ 
        Set $C(s,y)$ to 0.\\
    }
    
    \For{$k\gets 1$ \KwTo $M$} {
        \For{$y$ \rm{in} $\{0,1\}$} {
            $[\![LC\textit{Pr}(y)]\!]$ $\leftarrow$ $\pMUL([\![T(k)]\!],[\![LC(y,k)]\!])$\\ 
            $[\![C(p,y)]\!]$ $\leftarrow$ $[\![C(p,y)]\!] + [\![LC\textit{Pr}(y)]\!]$\\
            $[\![C(u,y)]\!]$ $\leftarrow$ $[\![C(u,y)]\!] + [\![LC(y,k)]\!] - [\![LC\textit{Pr}(y)]\!]$
        }    
    }
    
    \For{$s$ \rm{in} $\{p,u\}$ \rm{and} $y$ \rm{in} $\{0,1\}$} {
        $[\![C(s,y)]\!]$ $\leftarrow$ $[\![C(s,y)]\!]$ + \pLAP(1/$\epsilon$)\\
    }

    Set $N'$ to 0.\\
    \For{$s$ \rm{in} $\{p,u\}$ \rm{and} $y$ \rm{in} $\{0,1\}$} {
        $[\![N']\!]$ $\leftarrow$ $[\![N']\!] + [\![C(s,y)]\!]$\\ 
    }

    \For{$s$ \rm{in} $\{p,u\}$ \rm{and} $y$ \rm{in} $\{0,1\}$} {
         $[\![W(s,y)]\!]$ $\leftarrow$ \pDIV($[\![N']\!], 4 \cdot [\![C(s,y)]\!]$) \\
    }
   }
    \KwRet{ $[\![W(s,y)]\!]$}
    \caption{$\pPRE$ for privacy-preserving reweighing of training data \label{prot:preproc}}
\end{protocol}
\end{minipage}
\ \ 
\begin{minipage}[t]{0.49\textwidth}
\begin{protocol}[H]
   \SetKwInOut{Input}{Input}
    \SetKwInOut{Output}{Output}
    \Input{ Total number of samples as used for FL $N$,
    secret-shares of vectors $Y$, $\Hat{Y}$ and $S$ of length $N$
    }
    \Output{Secret-shares of confusion matrix for each group}
    {
       
       \For{$s$ \rm{in} $\{p,u\}$} {
            Set TP$(s)$, FP$(s)$, TN$(s)$, FN$(s)$ to 0 \\       
       }
       \For{$i\gets 1$ \KwTo $N$} {
       
       $[\![$tp$]\!]$ $\leftarrow$ $\pMUL([\![Y[i]]\!]$, $[\![\Hat{Y}[i]]\!]$) \\
       $[\![$ys$]\!]$ $\leftarrow$ $\pMUL([\![Y[i]]\!]$,  $[\![S[i]]\!]$) \\
       $[\![$ps$]\!]$ $\leftarrow$ $\pMUL([\![\Hat{Y}[i]]\!]$, $[\![S[i]]\!]$) \\
       $[\![$tps$]\!]$ $\leftarrow$ $\pMUL([\![$tp$]\!], [\![S[i]]\!]$) \\
       
       $[\![$TP$(p)]\!]$ $\leftarrow$ $[\![$TP$(p)]\!] + [\![$tps$]\!]$\\
       $[\![$TP$(u)]\!]$ $\leftarrow$ $[\![$TP$(u)]\!] + ([\![$tp$]\!] - [\![$tps$]\!] )$\\

       $[\![$FP$(p)]\!]$ $\leftarrow$ $[\![$FP$(p)]\!]+ ([\![$ps$]\!] - [\![$tps$]\!] )$\\
       $[\![$FP$(u)]\!]$ $\leftarrow$ $[\![$FP$(u)]\!] + ([\![\Hat{Y}[i]]\!]-[\![$ps$]\!]-[\![$tp$]\!] + [\![$tps$]\!] )$\\
       
       $[\![$FN$(p)]\!]$ $\leftarrow$ $[\![$FN$(p)]\!]+ ([\![$ys$]\!] - [\![$tps$]\!] )$\\
       $[\![$FN$(u)]\!]$ $\leftarrow$ $[\![$FN$(u)]\!] + ([\![Y[i]]\!]-[\![$ys$]\!]-[\![$tp$]\!] + [\![$tps$]\!] )$\\
       
       $[\![$TN$(p)]\!]$ $\leftarrow$ $[\![$TN$(p)]\!]+ ([\![S[i]]\!]-[\![$ys$]\!]- [\![$ps$]\!] + [\![$tps$]\!] )$\\
       $[\![$TN$(u)]\!]$ $\leftarrow$ $[\![$TN$(u)]\!] + (1 - [\![S[i]]\!] - [\![Y[i]]\!] + [\![$ys$]\!] - [\![\Hat{Y}[i]]\!] + [\![$ps$]\!] + [\![$tp$]\!] - [\![$tps$]\!])$\\
       }

    }
    \KwRet{ $([\![${\rm TP}$(s)]\!],[\![${\rm TN}$(s)]\!],[\![${\rm FP}$(s)]\!],[\![${\rm FN}$(s)]\!] \,|\, s \in \{u,p\})$}
    \caption{$\pCF$ for generation of confusion matrix for protected and unprotected group}\label{prot:pcf}
\end{protocol}
\end{minipage}

In Line 1--10 in Prot.~\ref{prot:preproc}, the computing parties compute secret shares of $C(s,y)$, for all values of $s$ and $y$, from the
secret shares of the clients' local counts $LC(y,k)$.
As is common in MPC protocols, we use multiplication instead of control flow logic for conditional assignments. In this way, the number and the kind of operations executed by the parties does not depend on the actual values of the inputs, so it does not leak information that could be exploited by side-channel attacks.
For instance, we rewrite ``$\textbf{if  } T(k) = 1 \textbf{  then  } C(p,y) \leftarrow C(p,y) + LC(y,k) \textbf{  else  } C(p,y) \leftarrow C(p,y)$'' as ``$C(p,y)  \gets  C(p,y) + T(k) \cdot LC(y,k) $'' (Line 6--7). 
Similarly, `$\textbf{if  } T(k) = 0 \textbf{  then  } C(u,y) \leftarrow C(u,y) + LC(y,k) \textbf{  else  } C(u,y) \leftarrow C(u,y)$'' is rewritten as  
``$C(u,y)  \gets  C(u,y) + (1-T(k)) \cdot LC(y,k) $''. The last expression can be rewritten as $C(u,y) + LC(y,k) - T(k) \cdot LC(y,k)$, allowing us to take advantage of the already computed multiplication in Line 6 to arrive at Line 8.
%
%
%
The counts $C(s,y)$ depend on disjoint subsets of the unified dataset $D$ (4 such disjoint subsets, given that the sensitive attribute and the class label are binary). The results of the counts can be made public under $\epsilon$-DP guarantee with the Laplace mechanism, i.e.~by adding Laplace noise with magnitude $1/\epsilon$ (noting that the sensitivity of each count query is 1, and that the parallel composition property of DP is applicable).
To add such noise, on Line 12, the parties call protocol $\pLAP$ to obtain a secret-shared value sampled from the Laplace distribution with mean 0 and median $1/\epsilon$. Pseudocode for $\pLAP$ is provided separately in Prot.~\ref{prot:lap} and explained below.
All the remaining computation in Prot.~\ref{prot:preproc} depend only on the now noisy counts. Because of DP's post-processing property, the outcomes of those computations also satisfy $\epsilon$-DP.
The computing parties then execute the MPC-protocol for division $\pDIV$ to compute secret shares of the weights on Line 19. The multiplication with constant $4$ ($4$ is the number of sensitive attribute values times the number of label values) and with $N'$ serve to rescale the weights so that they sum up to $N'$, which is the sum of the noisy counts $C(s,y)$ for all $s$ and $y$ (and likely differs from the real total number $N$ of training examples). 
At the end of the protocol, the computing parties hold secret shares of the computed weights protected under DP guarantees, which they then publish them so that the clients can use them for model training. $\pPRE$ not only protects the values of the clients' labels and sensitive attributes but also the value of the noise added to the computed weights.
\looseness=-1

Sampling from a Laplace distribution $Lap(0,b)$ with zero mean and scale $b$ in a privacy-preserving manner is achieved by letting the parties execute protocol \pLAP to compute a secret-sharing of  
$x = - b \cdot \sgn(u) \cdot \ln(1-2|u|)$, where $u$ is a random variable drawn from a uniform distribution in $[-0.5,0.5]$.\footnote{$x$ has distribution $Lap(0,b)$. This follows from the inverse cumulative distribution function for $Lap(0,b)$.} 
%
%
To obtain a secret-sharing of $u$, on Line 1 in in Prot.~\ref{prot:lap}, the parties execute \pRDM.
In \pRDM, each party generates $d$ random bits, where $d$ is the fractional precision of the power 2 ring representation of real numbers, and then the parties define the bitwise XOR of these $d$ bits as the binary representation of the random number jointly generated. The rest of $\pLAP$ is fairly straightforward. $\pLN$ called on Line 4 is a known MPC protocol for computing the natural logarithm of a secret-shared value \cite{mpspdz}. 


\textbf{Extending $\pPRE$ to other reweighing algorithms.}\label{SEC:preRWextend1}
Protocol $\pPRE$ can be easily extended to other reweighing techniques \cite{kamiran2012data} by computing the required statistics and adding noise to them before Line 14, and replacing Lines 14--20 to compute the weights according to the technique.  For example, for reweighing techniques that balance based only on $Y$
the parties would compute $C(y) = \sum_{s}C(s,y)$ after Line 17 (which is straightforward using addition of secret shares), which would be followed by a for-loop that iterates over $y$ and computes  $[\![W(y)]\!]$ $\leftarrow$ $\pDIV(N',2 \cdot [\![C(y)]\!])$.

\subsection{Privacy Preserving Post-processing for Bias Mitigation}\label{SEC:postproc}
In {\tt PrivFairFL-Post}, the predicted \textit{outcomes} are debiased by finding optimal classification thresholds for each group. 
In the centralized learning paradigm, this is done 
by constructing ROC-curves  
to find thresholds that maximize the given objective \cite{bird2020fairlearn,hardt2016equality}. 
This requires labels and predictions over the unified training data which violates privacy in the FL scenario.  
Below we detail a solution for obtaining ROC-curves and optimal classification thresholds in a privacy-preserving manner in FL. {\tt PrivFairFL-Post} is applied \textit{after} the training phase, so we assume that each client has obtained the last model $\theta$. Next:
\begin{enumerate}[noitemsep,topsep=0pt,leftmargin=*]
\item Each client $k$ generates predictions (probabilities) with $\theta$ over its local training dataset. Subsequently each client secret-shares with the MPC servers its sensitive attribute value, the predicted probabilities, and the ground truth labels for its local training examples.  
None of the MPC servers can derive any information about the true label, the sensitive attribute values and the predicted outcome from the encrypted shares received.
\item The MPC servers run protocol \pPOST (see Prot.~\ref{prot:postproc} to construct secret shares of the ROC curves for the protected and the unprotected group. Furthermore, noise is added inside the MPC protocol so that the ROC curves can be made public with $\epsilon$-DP guarantees.
\item All MPC servers send their secret shares of the noisy ROC curves to one of the MPC servers, who then computes optimal thresholds, and sends these to the clients. 
\end{enumerate}
\noindent
Step 1 is straightforward. We continue with describing steps 2 and 3. In 
protocol $\pPOST$, the parties construct ROC curves by computing secret-sharings of FPR and TPR values for a list of predefined candidate thresholds (see Line 1 in Prot.~\ref{prot:postproc} for the list). 
On Line 3--16, they compute secret shares of the $\mbox{TPR}$ and $\mbox{FPR}$ at each threshold, for the protected and the unprotected group.
On Lines 5--6, the parties use the comparison protocol $\pGTE$ to determine for each training instance ($i=1 \ldots N)$ whether it would be classified as positive or negative based on the $j$th threshold. This information, along with the ground truth labels $Y$ and the sensitive attribute values $S$ is then passed to a subprotocol $\pCF$ that returns secret-shares of TP, TN, FP and  FN for each group based on the $j$th threshold. Code for $\pCF$ is provided separately as Prot.~\ref{prot:pcf}; it is designed to minimize the amount of multiplications, and to avoid control flow statements (if-then-else) which would make the number of instructions performed dependent on the values of the input.
The sensitivity of each of the returned counts is 1, and they are based on disjoint subsets of the data $D$, so to provide $\epsilon$-DP, on Line 9--10 the parties draw secret-shared noise from $Lap(0,1/\epsilon)$ with $\pLAP$ (cfr.~Prot.~\ref{prot:lap}) and add it to the counts. 
The rest of the steps in Prot.~\ref{prot:postproc} are straightforward from the code.
All MPC servers then send their secret shares of the lists of FPRs, TPRs, and corresponding thresholds to one of the MPC servers who finds the optimal classification threshold for each group
following the procedure in \cite{bird2020fairlearn}. 

\begin{protocol}
   \SetKwInOut{Input}{Input}
    \SetKwInOut{Output}{Output}
    \Input{Total number of samples $N$ as used for FL ,
    vectors of secret-shares of true labels $Y$, predicted probabilities $\Hat{Y}_{prob}$ and $S$ each of length $N$;
    privacy budget $\epsilon$ allotted for bias mitigation}
    \Output{Secret-shares of ROC curves for protected and unprotected groups}
    {


       $[\![$thresholds$]\!]$ $\leftarrow$ $[\![(0.000, 0.001, 0.002, \ldots, 0.999, 1.000])\!]$; Set $T$ to 1001\\ 

       Declare $[\![$ROC$(p)]\!]$ and $[\![$ROC$(u)]\!]$ as secret-shared $3 \times T$ arrays.\\

       Declare $\Hat{Y}_{th}$ as an $N$-dimensional vector // \textit{holds predictions at thresholds}\\ 
       
       \For{$j\gets 1$ \KwTo $T$} {
       
       \For{$i\gets 1$ \KwTo $N$} {
          $[\![\Hat{Y}_{th}[i]]\!]$ $\leftarrow$ \pGTE($[\![\Hat{Y}_{prob}[i]]\!]$, $[\![$thresholds$[j]]\!]$)\\
       }
        
        $[\![$TP$(p)]\!],[\![$TN$(p)]\!],[\![$FP$(p)]\!],[\![$FN$(p)]\!], [\![$TP$(u)]\!],[\![$TN$(u)]\!],[\![$FP$(u)]\!],[\![$FN$(u)]\!] \leftarrow \pCF([\![Y]\!],[\![\Hat{Y}_{th}]\!],[\![S]\!])$\\
        

       \For{$s$ \rm{in} $\{p,u\}$}{
           $[\![$TP$(s)]\!]$ $\leftarrow$ $[\![$TP$(s)]\!]$ + \pLAP(1/$\epsilon$); $[\![$TN$(s)]\!]$ $\leftarrow$ $[\![$TN$(s)]\!]$ + \pLAP(1/$\epsilon$)\\
           $[\![$FP$(s)]\!]$ $\leftarrow$ $[\![$FP$(s)]\!]$ + \pLAP(1/$\epsilon$); $[\![$FN$(s)]\!]$ $\leftarrow$ $[\![$FN$(s)]\!]$ + \pLAP(1/$\epsilon$)\\

          $[\![$TPR$(s)]\!]$ $\leftarrow$ $\pDIV([\![$TP$(s)]\!],[\![$TP$(s)]\!] + [\![$FN$(s)]\!])$\\
          
           $[\![$FPR$(s)]\!]$ $\leftarrow$ $\pDIV([\![$FP$(s)]\!],[\![$FP$(s)]\!] + [\![$TN$(s)]\!])$\\
       
            $[\![$ROC$(s)[1,j]]\!]$ $\leftarrow$ $[\![$FPR$(s)]\!]$ \\
            
            $[\![$ROC$(s)[2,j]]\!]$ $\leftarrow$ $[\![$TPR$(s)]\!]$ \\
            
            $[\![$ROC$(s)[3,j]]\!]$ $\leftarrow$ $[\![$thresholds$[j]]\!]$ \\
       
       }
       }
 
    }
    \KwRet{ $[\![${\rm ROC}$(p)]\!]$,  $[\![${\rm ROC}$(u)]\!]$}
    \caption{$\pPOST$ for privacy-preserving debiasing of predicted outcomes}\label{prot:postproc}
\end{protocol}



\vspace{-3pt}
\subsection{Extensions to {\tt PrivFairFL}.}\label{SEC:extension}
\looseness=-1
%
Extending {\tt PrivFairFL} to cross-silo setups is straightforward. In {\tt PrivFairFL-Pre}, the clients will now instead secret-share local counts $LC(y,k,s)$ 
for \textit{each} value of $S$ (the count can be 0 if a certain sensitive attribute value is not present with the client), requiring only a small modification in the code for $\pPRE$ (Prot.~\ref{prot:preproc}). 
{\tt PrivFairFL-Post} can be used as is even in the cross-silo setup, as it is independent of the data distribution among the clients in FL. A straight-forward technique to extend our defined protocols for multi-class classification or multi-valued sensitive attributes is to adopt a one-vs-rest approach for each class and/or each value of sensitive attribute. This would then require the clients to release statistics for each binary combination of $S$ and $Y$ in a pre-defined sequence. Our approach can also be utilized in dynamic scenarios with client dropout and change in local data distributions, by computing the weights to be assigned to the training samples after a set of FL rounds. This will lead to a technique that is a combination of the pre-processing and in-processing techniques.
\vspace{-2pt}
\section{Results and Conclusion}\label{SEC:results}
\vspace{-3pt}
We evaluate {\tt PrivFairFL} on ADS~\cite{roffo2016personality} and ML-1M~\cite{MovieLens}. We implemented our MPC protocols for the pre-processing and the post-processing phase in MP-SPDZ \cite{mpspdz}. For the runtime experiments, every computing party ran on a separate VM instance co-located F48s V2 Azure virtual machines each of which contains 48 cores, 96 GiB of memory, and network bandwidth of up to 21 Gb/s. The implementations are all single-threaded.  All computations in Tab.~\ref{tab:mpcschemes} are done with $q=2^{64}$.
We implemented the model training phase in Python using Flower~\cite{beutel2020flower} for FL. We use FedAvg~\cite{mcmahan2017communication} as the aggregation strategy for the weights and add noise to the model parameters locally to provide DP guarantees during FL. In our experiments, we train some of the FL models with DP-SGD~\cite{abadi2016deep} as specified in Table~\ref{tab:results}.  We make all code, datasets and preprocessing pipelines publicly available to ensure reproducibility of our results. We provide a more detailed overview of the datasets, model architectures and hyperparameters in the appendix. 
We investigate the following research questions in our experiments:

\textbf{Q1: To what extent do our proposed pre-processing and post-processing approaches in {\tt PrivFairFL} improve fairness while addressing privacy?}
From the results in Tab.~\ref{tab:results}, it is seen that the fairness mitigation techniques aid in achieving group fairness in FL, and that the best type of technique depends on the dataset. {\tt PrivFairFL-Pre} benefits the highly imbalanced data in ADS, while {\tt PrivFairFL-Post} successfully finds optimal thresholds for the almost balanced subset of data in ML-1M. We also observe that randomization provided by DP guarantees can either aid in achieving a group-fair model or worsen the bias existing in the datasets. Our MPC+DP approaches not only aid in achieving group fairness while providing privacy guarantees but also improve the utility of the model when compared to the baseline approach using local DP (as done by \cite{abay2020mitigating}). We observe the fairness-utility trade-off in {\tt PrivFairFL-Pre} and {\tt PrivFairFL-Post}. 

\textbf{Q2: How does {\tt PrivFairFL} affect performance on real-world datasets?}
Based on our evaluation, {\tt PrivFairFL} can lead to less accurate models as opposed to models trained in CL and pure-FL. We think this is due to trade-offs for training fair models. Privacy-preserving techniques like DP can randomly affect the utility of the model, while our MPC+DP based approaches reduce such trade-offs and, as observed, improve utility over both the datasets.

\textbf{Q3: How does {\tt PrivFairFL} compare with existing work?} 
As per our evaluation, {\tt PrivFairFL} provides improvement over the baseline local DP approach based on \cite{abay2020mitigating}\footnote{adopted for cross-silo setup where each clients adds noise to their local counts and uses randomized response to publish the value of their sensitive attribute.} w.r.t.~utility and fairness. {\tt PrivFairFL} unlike the other techniques in Tab.~\ref{TAB:relatedwork}, not only reduce bias but also fully protects the data with provable privacy guarantees with little/no cost in performance.

\textbf{Q4: How does {\tt PrivFairFL} scale?}
We address the scalability of {\tt PrivFairFL} based on the number of clients and computing parties. Tab.~\ref{tab:mpcschemes} contains an overview of runtimes of our protocols for different MPC schemes with different computing parties and security settings. The runtime results show substantial difference between  passive and active security settings and between honest (3PC/4PC) and dishonest (2PC) majority settings; these results align with existing literature (\cite{dalskov2019secure, dalskov2021fantastic}). 
Our experiments are evaluated over a set of 75 and 109 clients, demonstrating the scalability of our approach. As could be expected, {\tt PrivFairFL-Post} takes much longer than {\tt PrivFairFL-Pre} to complete. We argue that even the longest runtimes of {\tt PrivFairFL-Post} are an acceptable price to pay for training fair models in a privacy-preserving way.

\begin{table}
    \caption{Runtimes for different MPC schemes for 2PC (dishonest majority) and 3PC/4PC (honest majority)} 
    \centering
    \scriptsize{
        \begin{tabular}{c c l rr rr}
        \toprule
        &  &   &  \multicolumn{2}{c}{ADS} & \multicolumn{2}{c}{ML-1M} \\
        \midrule
        &  & MPC scheme & {\tt Pre-proc.} & {\tt Post-proc.} & {\tt Pre-proc.} & {\tt Post-proc.}\\
        \midrule
        \multirow{2}{*}{passive}
        
        & 2PC &  OTSemi2k; semi-honest adapt.~of \cite{cryptoeprint:2018:482}&  0.10 sec & 43093.79 sec & 0.10 sec & 13088.52 sec\\
        
        & 3PC &  Replicated2k \cite{araki2016high}
        & 0.02 sec & 7993.73 sec & 0.02 sec & 2427.87 sec \\
                             
        \midrule
        \multirow{3}{*}{active}
        
        & 2PC &  SPDZ2k \cite{cryptoeprint:2018:482,damgaard2019new}   
        & 3.55 sec & 1199060.00 sec & 3.54 sec & 364180.50 sec\\
        
        & 3PC & SPDZ-wise Replicated2k \cite{dalskov2021fantastic}   
        & 0.06 sec & 16832.09 sec & 0.05 sec & 5112.27 sec\\
        
        & 4PC &  Rep4-2k \cite{dalskov2021fantastic}  
        & 0.02 sec & 8673.63 sec & 0.02 sec & 2634.37 sec\\ 
        
        \bottomrule
        \end{tabular} 
        }
    \label{tab:mpcschemes}
\end{table}





\begin{table}
  \caption{Utility and fairness for $\epsilon=1$ averaged over three runs with different value of seed }
  \label{tab:results}
  \centering
  \tiny{
  \begin{tabular}{lll  rrrrr  rrrrr}
    \toprule
    & Fairness & Privacy & \multicolumn{5}{c|}{\bf ADS }  & \multicolumn{5}{c}{\bf ML-1M}\\
    \midrule
    &    &   &
    Acc.   &  $|\mbox{1}$-$\mbox{DI}|$ & $\Delta$EOP & $\Delta$EODD & $\Delta$SP &   
    Acc.   &  $|\mbox{1}$-$\mbox{DI}|$ & $\Delta$EOP & $\Delta$EODD & $\Delta$SP  \\
    \midrule
    CL & $-$ & $-$ &
    85.81\%  & 1.214 & 0.070 & 0.037 & 0.004 &  
    62.15\%  & 0.138 & 0.091 & 0.141 & 0.094 \\
    
    FL & $-$ & $-$ &
    85.04\%  & 1.654 & 0.089 & 0.050 & 0.018 &  
    59.04\%  & 0.096 & 0.081 & 0.096 & 0.091 \\
    
    FL-DP-SGD & $-$ & DP-SGD &
    85.21\%  & 6.606 & 0.070 & 0.043 & 0.023 &  
    58.30\%  & 0.027 & 0.026 & 0.027 & 0.052 \\
    
    FL-DP-SGD & {\tt Pre} & Local DP &
    83.52\%  & 0.260 & 0.036 & 0.033  & 0.032 &  
    58.47\%  & 0.045 & 0.042 & 0.052 & 0.061 \\
    \midrule
    PrivFairFL  & {\tt Pre} & MPC+DP &
    83.57\% & 0.122  & 0.018 & 0.027 & 0.036  &  
    58.52\%  & 0.045 & 0.042 & 0.051 & 0.063  \\
    
    PrivFairFL  & {\tt Post} & MPC+DP &
    85.17\%  & 0.572 & 0.008 & 0.005 & 0.001 &  
    58.46\%  & 0.006 & 0.006 & 0.014 & 0.045 \\

    \bottomrule
  \end{tabular}
  }
\end{table}




\textbf{Conclusion \& Future Directions.} In this paper, we proposed {\tt PrivFairFL}, an MPC-based framework for training group-fair models in Federated Learning (FL). We proposed pre- and post-processing bias mitigation algorithms in cross-device FL under complete and formal privacy guarantees. We showed that {\tt PrivFairFL} not only efficiently balances the data and provides fair predictions, but also preserves the model quality. As a next step, we want to add in-processing bias mitigation techniques to {\tt PrivFairFL}. In addition to the group fairness notions that we have used in this paper, we plan to investigate other fairness notions, such as individual and causal notions of fairness, in our future work. 



\begin{ack}
Funding support for project activities has been partially provided by  the Microsoft Azure Research Credits Program, Canada CIFAR AI Chair and Facebook Research Award for Privacy Enhancing Technologies.
\end{ack}


\bibliography{references}
\bibliographystyle{plainnat}


\newpage
\appendix

\section{Limitations and Broader Impact}
In this work, we propose a framework to simultaneously achieve fairness and privacy in training a machine learning model in a federated scenario. However, it is important to note that our approach cannot work with all fairness notions. Since fairness notions are context dependent, it is clear that our framework cannot and should not be deployed for every application area. 

We acknowledge that our approach and our proposed protocols can be used in ways other than the ways mentioned in this paper. One can imagine that an attribute that should not be used to distinguish individuals could be used as a sensitive attribute, and the work presented in this paper can enable equalizing the performance across such groups in a private manner. Although our approach can produce fair and private predictions, it is still based on a model produced by a machine learning algorithm. Therefore, it becomes essential to understand that the {\tt PrivFairFL} model could also suffer from the same disadvantages as the original model in aspects that we didn't consider in this work, such as explainability, safety, security, and robustness. Hence, the user must be aware of such a system’s limitations, especially when using these models to replace people in decision making.

\section{Description of Protocol $\pCF$}
For protocol $\pCF$, the MPC servers receive as input secret-shares of a vector $Y$ with ground truth labels, secret-shares of a vector $\Hat{Y}$ with predicted labels, and secret-shares of a vector $S$ with sensitive attribute values. All vectors have length $N$, which is the total number of instances. The MPC servers execute protocol $\pCF$ to compute as output secret-shares of TP, TN, FP, and  FN for each group, i.e.~the protected group $p$ and the unprotected group $s$. We use the logic in the truth table shown in Table \ref{tab:truthtable} to compute the values of TP, FN, FP, and TN and replace control flow statements (if-then-else) by multiplications, as commonly done in MPC to make the number and kind of computations done during protocol execution independent of specific values of the data. Multiplying the columns in Table \ref{tab:truthtable} with the value $S[i]$ of the sensitive attribute as in the equations below, 
gives the contributions of instance $i$ to the TP, FN, FP, and TN scores of the protected and unprotected groups
\begin{equation*}
\mbox{TP}(p) = \sum\limits_{i=1}^N Y[i] \cdot \Hat{Y}[i] \cdot S[i]; \phantom{~~~~~~~~~~~~~~~~~~~~~~~~~~}\\
\mbox{TP}(u) = \sum\limits_{i=1}^N Y[i] \cdot \Hat{Y}[i] \cdot (1-S[i]); \\
\end{equation*}
\begin{equation*}
\mbox{FP}(p) = \sum\limits_{i=1}^N (1-Y[i]) \cdot \Hat{Y}[i] \cdot S[i];\phantom{~~~~~~~~~~~~~~~~~}\\
\mbox{FP}(u) = \sum\limits_{i=1}^N (1-Y[i]) \cdot \Hat{Y}[i] \cdot (1-S[i]); \\
\end{equation*}
\begin{equation*}
\mbox{FN}(p) = \sum\limits_{i=1}^N Y[i] \cdot (1-\Hat{Y}[i]) \cdot S[i]; \phantom{~~~~~~~~~~~~~~~~}\\
\mbox{FN}(u) = \sum\limits_{i=1}^N Y[i] \cdot (1-\Hat{Y}[i]) \cdot (1-S[i]); \\
\end{equation*}
\begin{equation*}
\mbox{TN}(p) = \sum\limits_{i=1}^N (1-Y[i]) \cdot (1-\Hat{Y}[i]) \cdot S[i]; \phantom{~~~~~~~}\\
\mbox{TN}(u) = \sum\limits_{i=1}^N (1-Y[i]) \cdot (1-\Hat{Y}[i]) \cdot (1-S[i]); \\
\end{equation*}

where $S[i] \in \{u,p\}$ is mapped to $S[i] \in \{0,1\}$ for mathematical computations ($u=0$ and $p=1$). 

We expand the above 8 equations and rewrite them as in Table \ref{tab:formula} by precomputing the common products
\begin{equation*}\label{products}
\mbox{tp} = Y[i] \cdot \Hat{Y}[i], \phantom{~~~~}   \\ 
\mbox{ys} = Y[i] \cdot S[i], \phantom{~~~~}          \\
\mbox{ps} = \Hat{Y}[i] \cdot S[i], \phantom{~~~~}    \\ 
\mbox{tps} = \mbox{tp} \cdot S[i]
\end{equation*}
for each instance to reduce the number of multiplications.

Following the above logic, the parties, on Lines 3--15 of protocol $\pCF$, compute the contribution of an instance towards  TP, FN, FP, or TN for the protected or unprotected group. On Lines 4--7, the parties precompute the secret shares of the common products (tp, ys, ps and tps) for each instance to reduce the number of multiplications. On Lines 8--15, the parties compute the secret shares of  TP, FN, FP, or TN based on the formulas in Table \ref{tab:formula} for the protected and unprotected groups.


%


\begin{table}
\centering
\caption{Logic for evaluating TP, FN, FP, and TN} 
\begin{tabular}{cc|cccc}
\toprule
& & true pos. & false neg. & false pos. & true neg.\\
$Y[i]$ & $\Hat{Y}[i]$  & $Y[i] \cdot \Hat{Y}[i]$ & $Y[i] \cdot (1-\Hat{Y}[i])$ & $(1-Y[i]) \cdot \Hat{Y}[i]$ & $(1-Y[i]) \cdot (1-\Hat{Y}[i])$ \\
\midrule
1 & 1 & 1 & 0 & 0 & 0 \\
1 & 0 & 0 & 1 & 0 & 0\\
0 & 1 & 0 & 0 & 1 & 0\\
0 & 0 & 0 & 0 & 0 & 1\\
\bottomrule
\end{tabular}
\label{tab:truthtable}
\end{table}

\section{Training Details}
We evaluated {\tt PrivFairFL} on two datasets, details of which are given below.

\textbf{ADS.} The ADS dataset\footnote{\url{https://www.kaggle.com/groffo/ads16-dataset}} is a collection of 300 advertisements explicitly rated by 120 users \cite{roffo2016personality}. Additionally, the dataset contains demographic information, interests, personality traits and textual phrases describing the likes and dislikes of each user. The advertisements are in the form of pictures. Each user rated each advertisement on a 1--5 scale. Following the categorization in \cite{roffo2016personality}, 
we labeled any ratings above 3 as ``positive'' and below and equal to 3 as ``negative''. 

We extracted features from the advertisements (number of faces, label types, safe category, objects) using Google Vision API.\footnote{\url{https://cloud.google.com/vision/docs/object-localizer}} 
The features for the users are based on one-hot-encoded information on their gender, country, interests, and income category, the first two indices of the zipcode, and their normalized age, along with the word embeddings\footnote{Based on pretrained Glove embeddings.} for the words describing the likes and dislikes of users.
Each training sample consists of the features of a user, features for an ad and the explicit rating provided by the user for the particular ad, leading to 300 samples for each user. We excluded all the users who had only negative ratings resulting in a total of 109 users and a total of 32700 training samples. 
We performed a stratified split over these users, keeping 80\% of their data as training data and the remaining 20\% data as the test data. 

To predict if a user is interested in an ad (``positive'' instance) we trained a dense layer with 2 units each with sigmoid activation. In the centralized setup (CL), this model is trained for 630 epochs with a batch size of 64. For the federated setup (FL), the models are trained for 800 rounds with each client training the local models for 2 epochs per round with a batch size of 24. For the models that we train with DP-SGD, the micro-batches are set same as the batch size with clipping threshold as 1 and noise multiplier as 0.7. We use SGD as the optimizer with glorot initializer and learning rate of 0.1, and binary cross-entropy as loss to train all the models. All the above models are trained thrice with different values of seeds (47568, 42, 1000), and the average values of the metrics are reported.

\begin{table}
\centering
\caption{Formulas for computing TP, FN, FP, and TN for protected and unprotected groups per instance} 
\begin{tabular}{c|cc}
\toprule
& Protected group & Unprotected group \\
\midrule
TP & tps & tp $-$ tps \\
FP & ps $-$ tps & $\Hat{Y}[i]$ $-$ ps $-$ tp $+$ tps \\
FN & ys $-$ tps & $Y[i]$ $-$ tp $-$ ys $+$ tps \\
TN & $S[i]$ $-$ ys $-$ ps $+$ tps  & 1 $-$ $S[i]$ $-$ $Y[i]$ $+$ ys $-$ $\Hat{Y}[i]$ $+$ ps $+$ tp $-$ tps  \\
\bottomrule
\end{tabular}
\label{tab:formula}
\end{table}


\textbf{ML-1K}
The Movielens dataset contains explicit 5-star movie ratings from 6040 users on 3952 movies. All users have rated at least 20 movies, but the count varies per user. Movie and user features are available, including demographic user information. The classification task is to predict whether a user would enjoy a particular movie. For this, we convert greater than 3 star reviews to a positive rating, and 3 stars or less to a negative rating, as with the ADS dataset. For movie features, we include one-hot-encoded genre and time between the user rating and the movie release year. For user features, we include categorical age, gender, and one-hot-encoded occupation. 

We take a sub-sample of 75 users for the experiments, within which the percentage of positive ratings is similar to the whole dataset (sub-sample: 58.11\%, whole dataset: 57.52\%), as is the percentage of female users (sub-sample: 32\%, whole dataset: 28.29\%). The model consists of a dense layer with 1 unit and sigmoid activation. In the centralized setting (CL), the model is trained for 20 epochs with a batch size of 32 and learning rate 0.015. In the federated setting (FL), the model is trained with 2 local epochs per user, 300 rounds of federated learning, with a learning rate of 0.03. The batch size and micro-batch size for DP-SGD are set to the size of the users' training set. The clipping threshold is set to 1.0 and the noise multiplier to 1.1. The model is trained with an SGD optimizer and binary cross-entropy loss. All results are reported on averages from 3 runs, using random seeds 1, 2 and 3. 

\textbf{Training with {\tt PrivFairFL}}
We recall that our proposed techniques for are independent of the model being trained. The model training times are thus the same as for training any DP-models in FL, and the cost for privacy-preserving debiasing with {\tt PrivFairFL} is separate.  Our experiments show that it takes about 32-40 minutes to train a LR model with DP-SGD on 109 clients with 800 federated runs. We also note that {\tt PrivFairFL-Post} is independent of any hyperparamters, giving the same convergence guarantees as state-of-the-art FL+DP training techniques. {\tt PrivFairFL-Pre} might benefit from additional rounds of FL or local epochs of the clients depending on the dataset. The number of rounds can be set by the aggregator in FL, who keeps track of the loss and utility on the held-out validation set and has knowledge on the kind of data. Additionally, we propose to use early stopping per client and setting a large maximum number of epochs. The convergence guarantees of {\tt PrivFairFL-Pre} thus also follow from the convergence guarantees of FL+DP.


\begin{figure}
  \centering
    \includegraphics[width=0.7\textwidth]{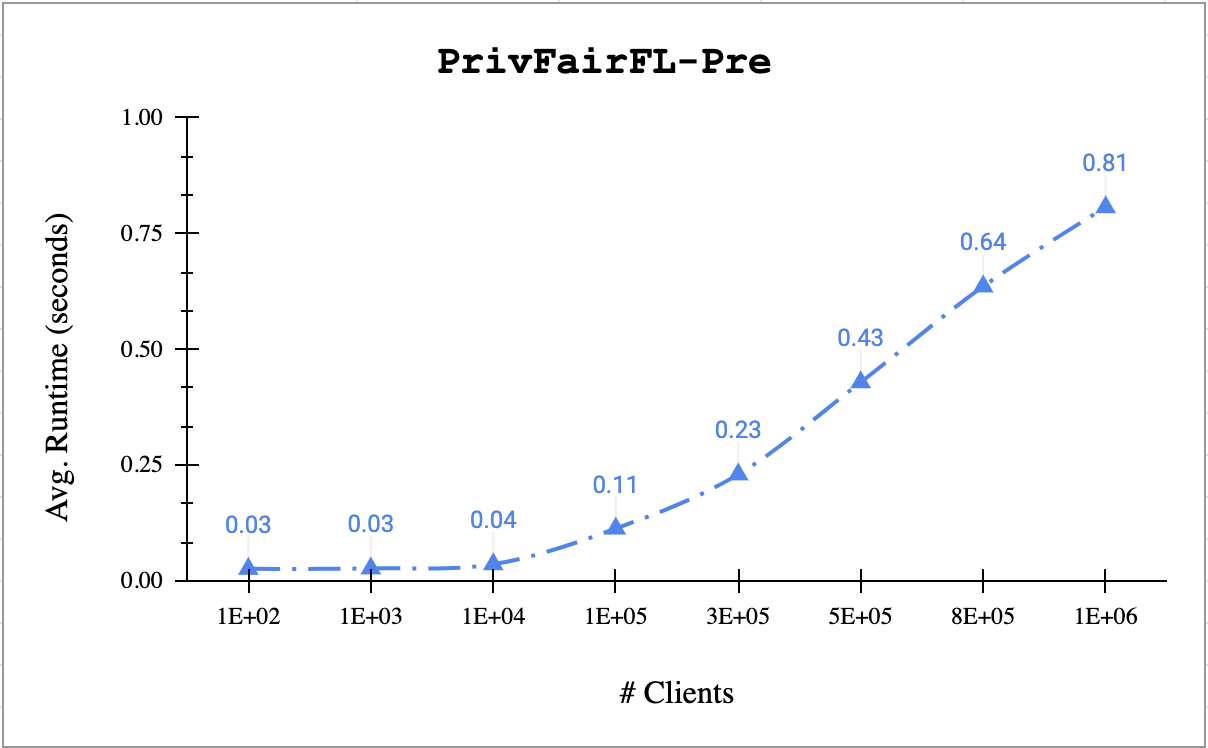}
  \caption{Runtimes for 3PC passive security setting for {\tt PrivFairFL-Pre}}
  \label{fig:clientsruntime-pre}
\end{figure}

\begin{figure}
  \centering
    \includegraphics[width=0.7\textwidth]{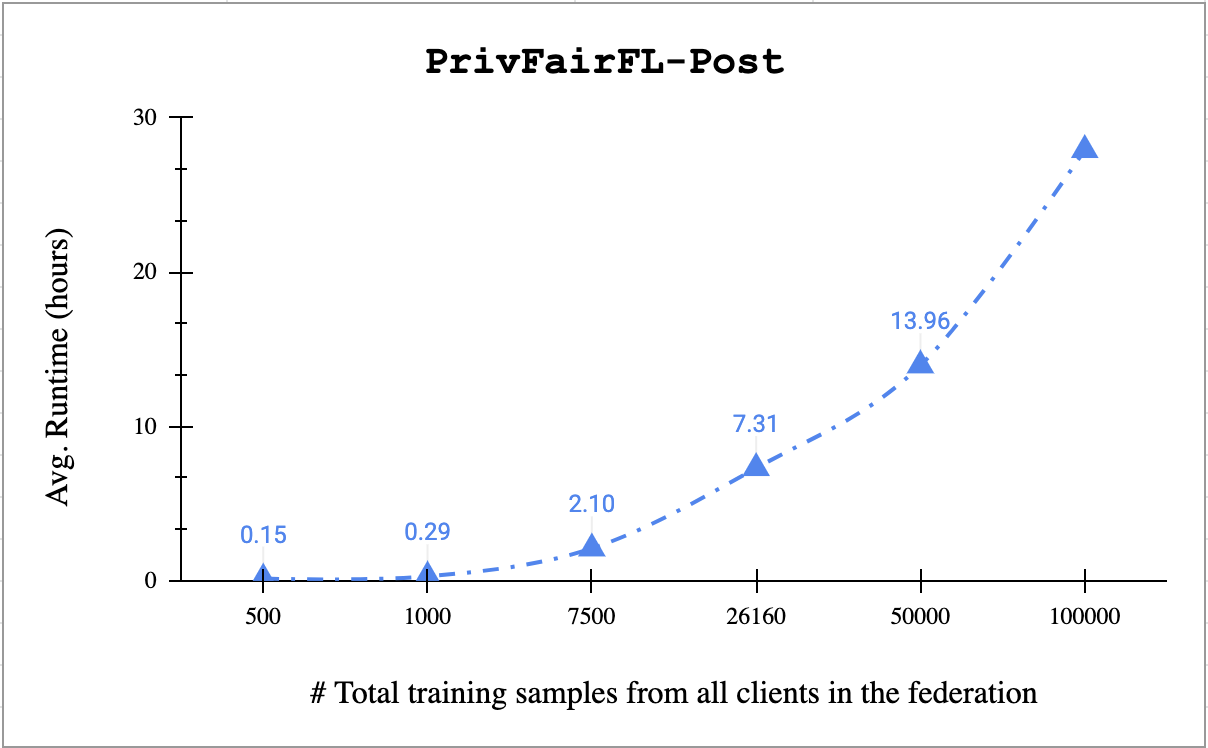}
  \caption{Runtimes for 3PC passive security setting for {\tt PrivFairFL-Post}}
  \label{fig:clientsruntime-post}
\end{figure}

\section{Additional Experiments}
Our contributions lie in achieving group fairness using simple yet effective bias mitigation techniques used in practice -- for imbalanced datasets and for biased predictions -- while preserving the privacy of the users. We ran experiments to evaluate the scalability of our proposed protocols and to study privacy-fairness-accuracy trade-offs. 

\paragraph{Scalability of proposed MPC protocols}
We provide complete and formal privacy guarantees by heavily relying on MPC protocols at the expense of longer runtimes. We note that our proposed techniques are independent of the training phase of the model and that the runtimes of the MPC protocols for bias mitigation are a one-time reasonable price to achieve both privacy and fairness.

{\tt PrivFairFL-Pre} essentially collects the statistics from each client and aggregates them to publish DP-weights for further federated learning. This is equivalent to collecting one vector from each client (one datapoint), making the aggregation for bias mitigation dependent only on the number of clients in the federation. Figure \ref{fig:clientsruntime-pre} shows the runtimes to execute $\pPRE$ with the number of clients in the federation ranging from 100 to 1,000,000, demonstrating that the runtimes increase linearly with the number of clients (note that the scale on the horizontal axis is logarithmic).\footnote{All the runtimes in Figure \ref{fig:clientsruntime-pre} and \ref{fig:clientsruntime-post} are averaged over 5 runs. We simulated the large number of clients and datapoints by making copies of the 109 clients and their training samples that we used for our experiments in the paper.} We see that for hundreds of thousands of clients, it takes about 0.8 seconds to debias the distributed dataset. While these runtimes may only roughly approximate what one could expect during actual deployment 
in dynamic environments, they indicate that our MPC protocol for preprocessing is practical enough to achieve both privacy and fairness. 



{\tt PrivFairFL-Post} collects the training labels, the inferred labels for the training data with the (unfair) model, and the sensitive attributes from each client in the federation to compute DP ROC curves   over all the training data from all the clients, where each client shares multiple datapoints. This makes our postprocessing technique dependent on the total number of training samples rather than the number of clients. Figure \ref{fig:clientsruntime-post} shows the runtimes to execute $\pPOST$ for an increasing number of datapoints, ranging from 10 to 100,000, considered for FL to compute the confusion matrices. We see that for a hundred thousand training samples contributed by clients in the federation, it takes about 28 hours to find the optimal classification thresholds for each sensitive group. We argue that this longer runtime could be acceptable as the thresholds need to be computed only once at the end of the federated model training. The participating clients then split their information (training labels, inferred labels for the training data with the learned model, sensitive attribute values) into secret shares and send them to the computing parties. The MPC overhead for the clients to contribute to model debiasing is negligible compared to the clients' model training costs. The MPC computational and communication overhead is instead carried by the computing parties who execute the MPC protocol, without requiring continued presence nor active participation of the clients.
The computing parties publish DP ROC curves to the FL aggregator who can then compute the thresholds to make the current model available for inference to the clients, at no additional cost to the clients. 


For practical deployment, the runtime of our solution can be further improved with optimizations of a cryptographic nature if we leave the MP-SPDZ framework. Secure multiplications are the bottleneck of any MPC-based solution. While we purposefully designed our MPC protocols to reduce the number of multiplications as much as possible, many multiplications remain. These secure multiplications are implemented using Beaver triples. A well known improvement is to pre-compute these triples during an ``offline phase'' (an option not available in MP-SPDZ) so that they are immediately available for consumption by the MPC protocols for postprocessing during the ``online phase''. The runtimes reported in our paper include both the pre-computations that can be done during an offline phase (before the computing parties become active) and an online phase (that starts when the clients send encrypted shares of their information to the computing parties). For ML tasks, the online phase is typically one to two orders of magnitude faster than the offline phase, i.e.~the duration of the online phase is only one tenth or less of the overall runtime \cite{mohassel2017secureml}.
Performing the offline computations prior to model debiasing (e.g.~during model training), can make our {\tt PrivFairFL-Post} more efficient for practical use.


\paragraph{Privacy-Utility-Fairness Trade-Offs of our proposed protocols}

The strength of our proposed techniques is using MPC which provides 
complete privacy to the \textit{inputs} while maintaining the utility. We note that MPC protocols are supported by mathematical proofs with no privacy loss metrics. 
We add noise to the aggregated \textit{outputs} using event level DP, which are then post-processed to compute outputs that see almost no variation in the ratio of the weights (PrivFairFL-Pre) and ratio of TPR and FPR (generated for PrivFair-Post). The simplicity of our approach is in secure aggregation of these statistics that aid in bias mitigation while preserving the utility of the model.
We can study the privacy and accuracy/fairness trade-offs only with respect to the privacy that we provide to the \textit{outputs} with respect to the privacy budget $\epsilon$ used in our MPC protocols to provide DP guarantees. With varying values of $\epsilon$ ranging from 0.25 to 50, we see almost no loss in utility or fairness, highlighting the advantages of emulated global DP using secure aggregation. For local DP on the other hand (i.e., MPC-free solution), for high values of $\epsilon$ ranging from 5 to 50, we did observe that the generated weights have large standard deviation, leading to a change of the ratios of the computed statistics that affect the fairness metrics. This is in line with known results that local DP affects utility much more aggressively.




\section{Code}
We have made the code available at \url{https://anonymous.4open.science/r/FairFL/}



\end{document}